\documentclass[runningheads]{llncs}

\usepackage{eccv}

\usepackage{eccvabbrv}

\usepackage{graphicx}
\usepackage{booktabs}
\usepackage{xspace}
\usepackage{array}
\usepackage{booktabs}
\usepackage{adjustbox}
\usepackage{multirow}
\usepackage[table]{xcolor}
\definecolor{first}{HTML}{B3CDE3} % Darker blue
\definecolor{second}{HTML}{CFE0ED} % Medium blue
\definecolor{third}{HTML}{EBF2F8} % Lighter blue
\usepackage{enumitem}
\usepackage{ragged2e} % For \RaggedRight
\usepackage{arydshln} % For dashed lines (\hdashline)
\renewcommand{\arraystretch}{1.2} % Adjust row spacing
\usepackage{lipsum}   % For dummy text (optional)
\usepackage{algorithm}
\usepackage{algorithmic}
\usepackage{amsmath}
\usepackage{amssymb}
\usepackage{pifont}
\usepackage{subcaption}
\usepackage{caption}   % required for \captionof
\usepackage{float}   % optional, for [H]
\usepackage{enumerate}
\usepackage{enumitem}

\newcommand{\checked}{\ding{51}}
\newcommand{\crossed}{\ding{55}}

\usepackage[accsupp]{axessibility}  % Improves PDF readability for those with disabilities.

\usepackage[breaklinks,colorlinks,citecolor=eccvblue]{hyperref}
\usepackage{orcidlink}

\begin{document}

% ---------------------------------------------------------------
% TODO REVIEW: Replace with your title
\title{MarineEVT: Advancing Event-Centric Marine Video Understanding via Visual Tool Reasoning} 

% MarineEVT: Enhancing and Benchmarking VLMs for Marine Event-Centric Understanding via Episodic Visual Tool Invocation

% 往tool design和调用方向去写？往agent调用方向去体现dataset construction的能力？

% Write a pseudo code to describe your agentic flow.

% TODO REVIEW: If the paper title is too long for the running head, you can set
% an abbreviated paper title here. If not, comment out.
\titlerunning{MarineEVT}

% TODO FINAL: Replace with your author list. 
% Include the authors' OCRID for the camera-ready version, if at all possible.
\author{Tuan-An To\inst{1}\orcidlink{0000-0002-5408-8138} \and
Yuk-Kwan Wong\inst{1}\orcidlink{0009-0007-4444-5502} \and
Tuan-Anh Vu\inst{2}\orcidlink{0000-0002-8872-0875}
\and \\
Ziqiang Zheng$^{\dagger}$\inst{1,3}\orcidlink{0000-0002-1477-6040}
\and 
Sai-Kit Yeung\inst{1}\orcidlink{0000-0001-7974-0607}
\\
\small
Project website: \url{https://marineevt.hkustvgd.com}; \quad $\dagger$~: zhengziqiang1@gmail.com
}

% TODO FINAL: Replace with an abbreviated list of authors.
\authorrunning{A. To et al.}
% First names are abbreviated in the running head.
% If there are more than two authors, 'et al.' is used.

% TODO FINAL: Replace with your institution list.
\institute{The Hong Kong University of Science and Technology, Hong Kong, China \and
University of California, Los Angeles
CA, USA \and
University of Electronic Science and Technology of China, China}

\maketitle

\begin{center}
   \includegraphics[width=\textwidth]{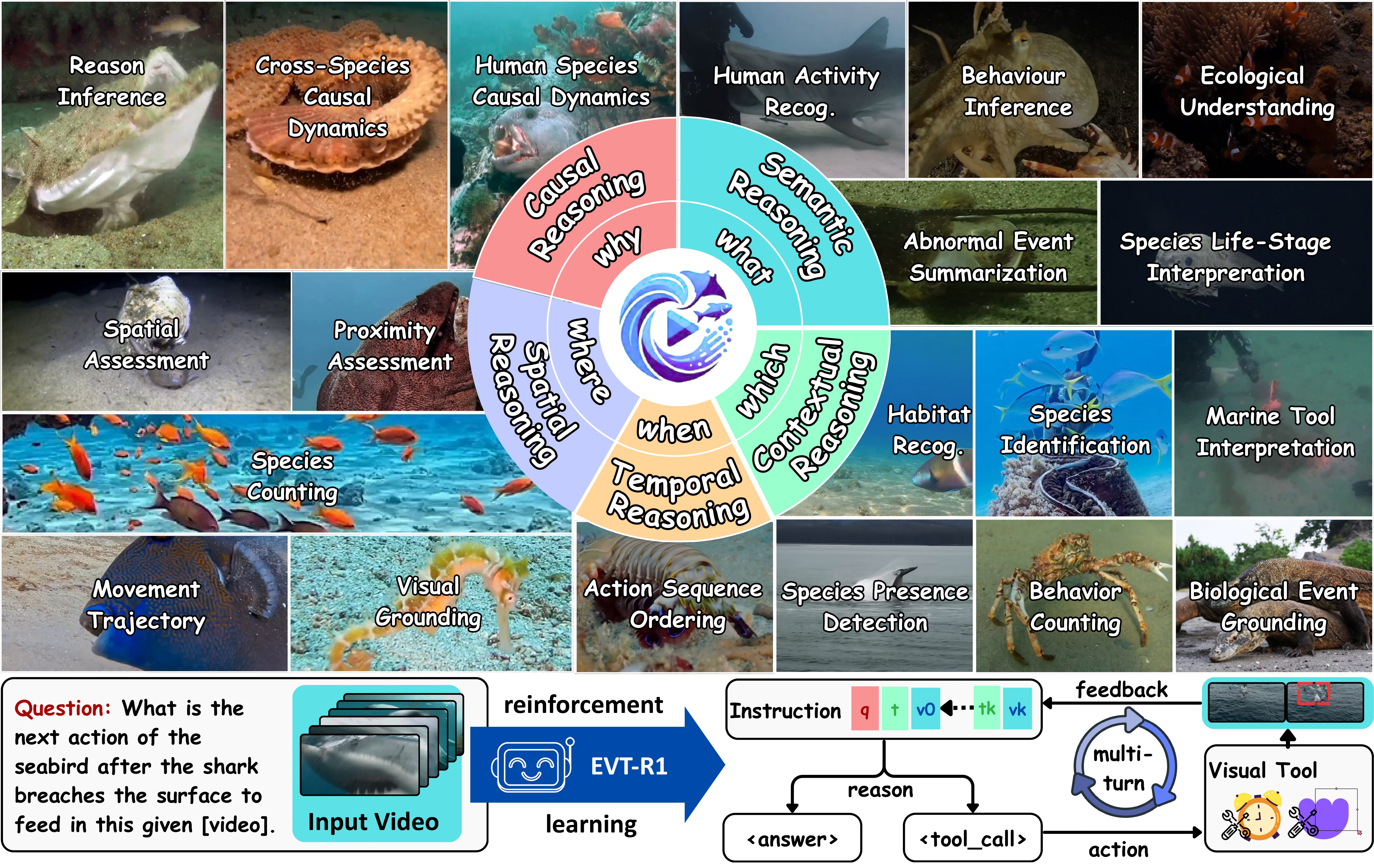}
   \captionsetup{type=figure}
   \vspace{-0.25in}
   \captionof{figure}{We propose MarineEVT, the first hierarchical and comprehensive event-centric marine video dataset. Based on MarineEVT, we propose EVT-R1, integrating visual tool reasoning into the VLM for more reliable marine video understanding. \label{fig:teaser}}
   \vspace{-0.15in}
\end{center}

\begin{abstract}
Recent Vision-Language Models (VLMs) have achieved remarkable success in visual understanding, driven by the growing availability of high-quality image-text pairs. However, the performance of VLMs often degrades in the video domain due to the essential need for temporal understanding and the scarcity of large-scale annotated video data. In this work, we focus on marine video understanding, which brings further challenges: first, it requires substantial domain expertise; and video VLMs usually struggle with localizing and interpreting critical information from marine videos, as the informative events are typically \textit{sparse}, \textit{unpredictable}, and \textit{unevenly distributed}. To address these challenges, we carefully curate the first event-centric marine video understanding dataset called \textbf{MarineEVT}, which features 20K multi-task, video-level visual question-answering pairs spanning multiple dimensions of marine understanding and analysis. Meanwhile, based on MarineEVT, we decompose marine video understanding as an Event-centric Visual Tool-integrated Reasoning process (\textbf{EVT-R1} for short), where we leverage powerful visual tools to drive the model to localize and interpret critical information aligned with visual questions and human intent. To demonstrate its effectiveness, we compare EVT-R1 against 11 SOTA VLMs in different settings. EVT-R1 outperforms the top open-source and top commercial models by 5.22 and 11.09, respectively. MarineEVT and EVT‑R1 lay the foundation for ecological discovery and marine education, fostering the development of VLMs capable of interpreting marine dynamics, reasoning about ecological interactions, and supporting sustainable ocean video understanding and analysis. % \href{https://huggingface.co/datasets/AnTo2209/MarineEVT}{Explore Dataset Here}
\end{abstract}
\vspace{-0.15in}

\section{Introduction}
\label{sec:intro}
Marine understanding stands as a pivotal frontier in biological and environmental science, shaping our ability to study and preserve the vast, complex ecosystems that cover over 70\% of Earth yet remain little explored. Uncovering its secrets~\cite{xue2025uvlmbenchmarkingvideolanguage,zhang2025uwbenchcomprehensivevisionlanguagebenchmark,zhang2024webuot1madvancingdeepunderwater,zhang2024fantasticanimalsthemsegment} is crucial for both advancing scientific understanding and tackling global challenges such as biodiversity loss~\cite{hughes2018large}, climate change~\cite{zhong2023combining}, and sustainable ocean management~\cite{winther2020integrated}. In recent years, this domain has experienced remarkable advancements, particularly in single-image analysis~\cite{ziqiang2024marineinst,uveb2024}, where state-of-the-art methods have demonstrated outstanding performance across a broad spectrum of visual tasks, including marine object detection~\cite{fan2020dual,wong2025orca,wille2025all}, instance segmentation~\cite{lian2023watermask}, instance-level captioning~\cite{ziqiang2024marineinst}, and related applications~\cite{hong2023usod10k,uveb2024}. 

However, despite these remarkable advances in image-level analysis, progress in comprehensive marine video understanding remains considerably constrained~\cite{uveb2024,xue2025uvlmbenchmarkingvideolanguage}. Unlike image-level visual understanding~\cite{zheng2023marinegpt,zhang2025uwbenchcomprehensivevisionlanguagebenchmark}, video understanding is inherently more challenging, as it demands \textbf{event-level interpretation}. The complex intrinsic of marine exploration challenges effective marine video understanding: 1) marine videos are often \textbf{long} and \textbf{tedious}, making it difficult to localize and interpret meaningful events; 2) comprehending the marine videos (especially videos with ecological traits) requires \textbf{significant domain expertise}; 3) marine video understanding tailored to ecological and educational purposes sets it apart from general-purpose applications.

In marine videos, informative events are often \textit{sparse}, \textit{ephemeral}, and \textit{unevenly distributed}, posing significant challenges for existing VLMs~\cite{bai2025qwenvl3,bai2025qwenvl2.5,bai2024qwenvl2} to effectively localize and interpret these sparse but crucial events. These episodic observations carry immense scientific value~\cite{gonzalez2023survey, oceancv2024}, offering critical insights for ecological dynamics~\cite{clibd2024, biodiversity2023}, species behavior~\cite{marine2025,jalal2023fish}, biodiversity evolution~\cite{hughes2018large}, and the impacts~\cite{frontiers2026ml, zhong2023combining} of human activities or climate change. However, current general-purpose VLMs primarily emphasize scene summarization, overlooking the fine-grained visual dynamics that better align with domain requirements. Thus, marine video understanding demands a significant shift toward precise, event-centric descriptions that capture specific entities (\emph{e.g.}, marine organisms, divers, instruments) and their subtle and dynamic interactions. This gap underscores the need for specialized, domain‑adaptive video VLMs capable of capturing informative visual dynamics for event‑centric marine video understanding.

Motivated by such demand, we carefully construct an event-centric dataset and benchmark, \textbf{MarineEVT}, as shown in Fig.~\ref{fig:teaser}, to drive VLMs toward deeper temporal reasoning and domain‑aware understanding of dynamic marine events. MarineEVT comprises 20,000 richly annotated underwater video question-answer pairs spanning 20 fine-grained dimensions, including \textit{marine species}, \textit{human activities}, \textit{environmental conditions}, \textit{behavioral interactions}, and \textit{rare ecological events}, structured to support \textbf{semantic}, \textbf{contextualized}, \textbf{spatial-temporal}, and \textbf{causal} reasoning. During the construction of MarineEVT, we integrate marine domain expertise into \textit{prompt design} and \textit{annotation verification}, ensuring ecologically precise, context‑aligned supervision that helps VLMs interpret biologically meaningful cues and minimize ambiguity. Lastly, unlike existing marine datasets~\cite{zheng2023marinegpt, ziqiang2024marineinst, wong2025marineeval, xue2025uvlmbenchmarkingvideolanguage} that focus on global scene description, our dataset centers on localizing and understanding meaningful events. 

\begin{figure}[t]
    \centering
   \includegraphics[width=\textwidth]{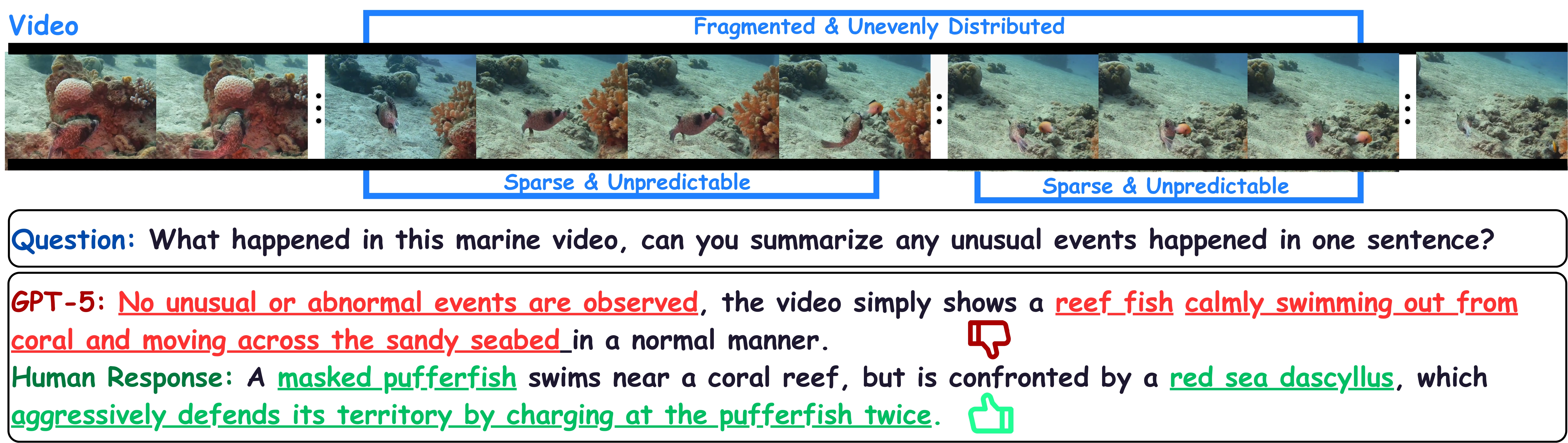}
   % \vspace{-0.25in}
   \captionof{figure}{An example question for evaluating event summarization tested on GPT-5.0~\cite{singh2025openaigpt5card} with human response provided for comparison. 
    } % Supporting footage \href{https://www.instagram.com/p/CwFqYt7Nlq2/}{here}
   \label{fig:motivation}
   \vspace{-0.25in}
\end{figure}

Although MarineEVT fulfills the need for event-centric marine video datasets, we observe that general-purpose VLMs struggle with limited domain expertise and an inability to localize or retrieve critical information aligned with visual questions and human intent, as shown in Fig.~\ref{fig:motivation}. The advanced models like GPT-5~\cite{singh2025openaigpt5card} fail to accurately summarize the happening event, a reef fish defending territory, highlighting the ecological significance. The abundance of redundant visual inputs across video, caused by slow or static underwater motion, obscures the boundaries of meaningful events and weakens the interest of frames that favor the domain requirements. These weaknesses particularly hinder the model's ability to ground language to sparse, fragmented events, such as predator–prey interactions or rare species appearances, which are ecologically significant yet easily lost in the episodic videos. Even fine-tuning general-purpose VLMs on MarineEVT can enrich their domain knowledge, the fine-tuned VLMs will still struggle with robust temporal reasoning and event-level interpretation, since there is no specific design to localize the sparse but critical events from marine videos with redundant frames. 

In this work, we propose decomposing complex event-centric marine video understanding into \textbf{consecutive reasoning turns}. Our motivation is aligned with how humans solve complicated tasks (especially for the tasks that require deep expertise): they conduct tasks in multiple turns, in each turn they decide whether to use external tools for assistance and the intermediate outputs produced serve as extra information for the final outputs. Specifically, we leverage powerful visual tools to retrieve and localize critical information from redundant visual inputs. Meanwhile, the tool invocation also yields extra visual cues/guidance for condensing visual representations and driving the VLM towards more reliable answer generation. The proposed Event-centric Visual Tool-integrated Reasoning framework (\textbf{EVT-R1} for short) enables the VLM to focus on critical events and relevant entities, discarding irrelevant visual signals to enhance temporal grounding and dynamic context understanding. 

Equipped with EVT-R1, we achieve better performance gains (\textcolor{green}{+5.22}) than supervised fine-tuning (\textcolor{green}{+1.6}), and also reinforcement learning (\textcolor{red}{-8.09}) for post-training only. Furthermore, we benchmark 11 SOTA models across a comprehensive suite of tasks, including multi‑task spatial/temporal grounding, video question answering, and video summarization, under different settings. The experimental analysis provides insights regarding developing domain-adaptive VLMs and sparks a new direction for integrating visual tools for advancing event‑centric marine video‑language understanding. The main contributions of this paper are summarized as follows: 

\begin{itemize}[leftmargin=0pt]
\item We curate \textbf{MarineEVT}, a dataset of 20K video question-answer pairs spanning 20 distinct dimensions and reasoning tasks. To the best of our knowledge, this is the first dataset and benchmark for event-centric marine video understanding.
\item We propose decomposing event-centric marine video understanding into a multi-turn \textbf{visual tool-integrated reasoning} process, leveraging powerful visual tools to localize and interpret critical information from redundant video frames with sparse and unevenly distributed events.
\item We propose \textbf{EVT-R1}, a training paradigm that equips VLMs with multi‑turn, event‑grounded reasoning capabilities to better process complex marine video dynamics. It devises dual rewards to optimize tool use and answer generation, yielding more robust and interpretable domain-specific understanding.
\end{itemize}

\section{Related Work}
\label{sec:related_work}
% \vspace{-0.08in}
\subsection{Event Understanding}
% \vspace{-0.06in}
% While marine videos are increasingly abundant, fueled by the exponential growth of online video content, their effective utilization is hindered by major challenges
Event understanding~\cite{jang2023knowing,sanders2024survey,ramanathan2013video} involves comprehending how visual scenes evolve over time by interpreting interactions, transitions, and causal relationships between entities within a dynamic context~\cite{tang2025videounderstandinglargelanguage, feichtenhofer2019slowfastnetworksvideorecognition}. It extends beyond static visual perception to encompass the recognition~\cite{simonyan2014twostreamconvolutionalnetworksaction}, identification~\cite{activitynet2025cvpr}, and causal reasoning~\cite{causal2023cvpr} of actions or scene changes over time. Through event understanding, we can integrate spatial, temporal, and contextual information to form a coherent understanding of what is happening, why it occurs, and what may happen next. To achieve this, recent multimodal frameworks propose to incorporate linguistic cues to promote interpretability, such as VideoCLIP~\cite{xu2021videoclip}, MERLOT~\cite{rowan2021merlot}, InternVL series~\cite{chen2024internvl, chen2024internvl2, zhu2025internvl3, wang2025internvl3_5}, and QwenVL series~\cite{bai2023qwenvl, bai2024qwenvl2, bai2025qwenvl2.5, bai2025qwenvl3}, which integrate visual dynamics with textual guidance to model temporal relationships and narrative semantics. Despite their effectiveness, these models heavily rely on large-scale pretraining corpora~\cite{wang2023internvid,wang2024event} and often overlook fine-grained, domain-specific events (\emph{e.g.}, surgical actions~\cite{kim2025surgical}, scientific procedures~\cite{zhao2025mmvu}, or ocean-monitoring~\cite{shi2022detecting} workflows), where data distributions and semantics differ substantially from those of internet-scale datasets. In detail, our daily activity videos~\cite{activitynet2025cvpr, damen2018scalingegocentricvisionepickitchens,han2024videoespressolargescalechainofthoughtdataset} are typically structured, predictable, and even summarizable as a series of steps or procedures. In contrast, some domain-specific videos, such as surgical procedures~\cite{zeng2025surgvlmlargevisionlanguagemodel, li2025surgpubvideocomprehensivesurgicalvideo} or underwater exploration~\cite{xue2025uvlmbenchmarkingvideolanguage} recordings, exhibit complex and less predictable dynamic events. 

To capture these dynamics, recent VLMs have incorporated event-centric vision encoders that enhance spatial-temporal representation learning. EventGPT~\cite{liu2025eventgpt} and EventVL \cite{li2025eventvl} explicitly encode event structure into the visual backbone to improve temporal coherence. Reinforcement learning~\cite{murphy2025reinforcementlearningoverview} provides a pathway toward interpretable, event-aware modeling. Recent advances~\cite{rafailov2023direct, guo2025grpo, yu2025dapoopensourcellmreinforcement, zheng2025groupsequencepolicyoptimization} enable LLMs to perform action-level reasoning under policy and reward constraints, fostering deliberate, stepwise inference. Specifically, VTool-R1 \cite{wu2025vtoolr1vlmslearnthink} employs reinforcement learning to fine-tune VLMs to explore flexible reasoning trajectories and learn to use visual editing tools effectively. Yet this paradigm remains underexplored in marine video understanding, where agents must recognize salient moments, infer causal interactions, and reason about underwater events as they explore. 

% Glance-Focus~\cite{bai2023glance} introduces a memory mechanism with a memory queue to retain historical context and enable temporally grounded question answering. SEG~\cite{dixit2026semanticeventgraphslongform} constructs temporal scene graphs with relational prompts to improve event understanding. However, existing models depend on implicit representations, overlooking explicit reasoning about key frames and causal dynamics. 

% \vspace{-0.12in}
\subsection{Marine Understanding}
% \vspace{-0.06in}
Marine datasets and benchmarks have progressed rapidly, comprehensively addressing tasks such as instance segmentation~\cite{lian2023watermask, muk2025uod}, object detection~\cite{hong2023usod10k, muk2025uod,wong2025orca}, and object tracking~\cite{zhang2024webuot1madvancingdeepunderwater, alawode2022utb180}. However, perceptually complex tasks requiring semantic reasoning, contextual understanding, and domain-specific knowledge in vision-language understanding remain underexplored. MarineGPT~\cite{zheng2023marinegpt} and MarineInst~\cite{ziqiang2024marineinst} introduce multimodal benchmarks that combine visual and linguistic understanding of marine imagery. More recently, CoralVQA~\cite{han2025coralvqa} introduced a large-scale dataset specialized for coral reef understanding. MarineEval~\cite{wong2025marineeval} introduced a multi-task framework evaluating marine intelligence across VQA, summarization, grounding, and completion for VLMs. These benchmarks remain constrained to static image-level assessment, overlooking the rich temporal dynamics. Because marine observations unfold through temporally and causally linked events, their interpretation demands domain expertise beyond static-image analysis; event-centric understanding is crucial for capturing ecological dynamics. Although UVLM~\cite{xue2025uvlmbenchmarkingvideolanguage} initiated marine video analysis, it remains limited to static attributes, such as 
species labels or isolated actions, overlooking the temporal and causal dynamics central to event-level video understanding explored in our work. 

% To bridge this domain gap, recent work has adapted VLMs to specialized contexts through tailored alignment strategies. Examples include MedCLIP \cite{wang2022medclip} for medical imaging, GeoDANO\cite{cho2025GeoDANOGV} for geometric reasoning as well as DriVLM and DriveLM~\cite{zheng2025drivl,sima2023drivelm} for autonomous driving scenarios, which refine representation bottlenecks and alignment objectives to accommodate task-specific semantics. These efforts show that effective transfer of visual-language priors to event-centric domains requires explicit modeling of temporal dynamics and causal coherence, not just solely fine-tuning.
% InternVideo~\cite{wang2022internvideo, wang2024internvideo2, wang2025internvideo} and VideoLLaMA~\cite{damonlpsg2023videollama, damonlpsg2024videollama2, damonlpsg2025videollama3} 

% Within domain-specific settings, \textit{event} understanding necessitates modeling not only entities and their interactions but also the temporal dependencies that govern their evolution. 
\begin{table}[t]
\centering
\caption{\small Comparisons with existing marine (VLM \& non-VLM) datasets.\label{tab:comparisions}
 }
 \vspace{-4mm}
 \protect \begin{tikzpicture}[baseline=(char.base)]
        \node[shape=circle, draw=black, fill=orange!50, scale=0.55, inner sep=1pt] (char) {\textbf{A}};
    \end{tikzpicture}: Attribute,
    \begin{tikzpicture}[baseline=(char.base)]
        \node[shape=circle, draw=black, fill=purple!40, scale=0.56, inner sep=1pt] (char) {\textbf{B}};
    \end{tikzpicture}: Behaviour,
    \begin{tikzpicture}[baseline=(char.base)]
        \node[shape=circle, draw=black, fill=red!20, scale=0.65, inner sep=1pt] (char) {\textbf{S}};
    \end{tikzpicture}: Species,
    \begin{tikzpicture}[baseline=(char.base)]
        \node[shape=circle, draw=black, fill=yellow!20, scale=0.6, inner sep=1pt] (char) {\textbf{H}};
    \end{tikzpicture}: Human,
    \begin{tikzpicture}[baseline=(char.base)]
        \node[shape=circle, draw=black, fill=green!20, scale=0.6, inner sep=1pt] (char) {\textbf{E}};
    \end{tikzpicture}: Environment, \\
    \begin{tikzpicture}[baseline=(char.base)]
        \node[shape=circle, draw=black, fill=white!20, scale=0.65, inner sep=1pt] (char) {\textbf{S}};
    \end{tikzpicture}: Static,
    \begin{tikzpicture}[baseline=(char.base)]
        \node[shape=circle, draw=black, fill=gray!30, scale =0.6, inner sep=1pt] (char) {\textbf{D}};
    \end{tikzpicture}: Dynamic,
    \begin{tikzpicture}[baseline=(char.base)]
        \node[shape=circle, draw=black, fill=pink!50, scale=0.6, inner sep=1pt] (char) {\textbf{T}};
    \end{tikzpicture}: Temporal,
    \begin{tikzpicture}[baseline=(char.base)]
        \node[shape=circle, draw=black, fill=blue!30, scale=0.6, inner sep=1pt] (char) {\small \textbf{R}};
    \end{tikzpicture}: Reason, \begin{tikzpicture}[baseline=(char.base)]
        \node[shape=circle, draw=black, fill=blue!50, scale=0.6, inner sep=1pt] (char) {\small \textbf{O}};
    \end{tikzpicture}: Outcome.
\vspace{-2mm}
\small % Smallest standard font size
\setlength{\tabcolsep}{4pt} % Very tight column padding
\resizebox{\columnwidth}{!}{%
    \begin{tabular}{lcccccc|cccccccc}
    \toprule
    \textbf{Dataset} & \textbf{V. Modality} & \textbf{Q. Format} & \textbf{Dataset Task} & \textbf{\#Data} & \textbf{\#Dimension} & \textbf{Tool} & \textbf{Semantic.} & \textbf{Contextual.} & \textbf{Spatial.} & \textbf{Temporal.} & \textbf{Causal.} \\
    \midrule \midrule
    % SUIM~\cite{islam2020suim}
    % & Image & masks & Segmentation & 1.5K & $-$ 
    % & \crossed & \crossed & \crossed  & \checked  & \crossed  & \crossed \\
    UIIS~\cite{lian2023watermask} & Image & masks & Segmentation & 5K  & $-$ 
    & \crossed & $-$ & \begin{tikzpicture}[baseline=(char.base)]
        \node[shape=circle, draw=black, fill=red!20, inner sep=1pt] (char) {\textbf{S}};
    \end{tikzpicture}  & \begin{tikzpicture}[baseline=(char.base)]
        \node[shape=circle, draw=black, fill=white!20, inner sep=1pt] (char) {\textbf{S}};
    \end{tikzpicture}  & $-$ & $-$ \\
   FishNet~\cite{khan2023fishnet} & Image  & labels/bounding-boxes & Recog./Detect. & 95K  & $-$ & \crossed & $-$ & \begin{tikzpicture}[baseline=(char.base)]
        \node[shape=circle, draw=black, fill=red!20, inner sep=1pt] (char) {\textbf{S}};
    \end{tikzpicture}  & \begin{tikzpicture}[baseline=(char.base)]
        \node[shape=circle, draw=black, fill=white!20, inner sep=1pt] (char) {\textbf{S}};
    \end{tikzpicture}  & $-$ & $-$ \\
   Ocean20K~\cite{li2025exploring} & Image  & masks & Segmentation & 20K  & $-$ 
   & \crossed & $-$ & \begin{tikzpicture}[baseline=(char.base)]
        \node[shape=circle, draw=black, fill=red!20, inner sep=1pt] (char) {\textbf{S}};
    \end{tikzpicture}  & \begin{tikzpicture}[baseline=(char.base)]
        \node[shape=circle, draw=black, fill=white!20, inner sep=1pt] (char) {\textbf{S}};
    \end{tikzpicture}  & $-$ & $-$ \\
   MarineInst~\cite{ziqiang2024marineinst} & Image & masks/open-end & Seg./Caption. & 20M & $-$ 
    & \crossed & \begin{tikzpicture}[baseline=(char.base)]
        \node[shape=circle, draw=black, fill=orange!50, inner sep=1pt] (char) {\textbf{A}};
    \end{tikzpicture} & \begin{tikzpicture}[baseline=(char.base)]
        \node[shape=circle, draw=black, fill=red!20, inner sep=1pt] (char) {\textbf{S}};
    \end{tikzpicture}  & \begin{tikzpicture}[baseline=(char.base)]
        \node[shape=circle, draw=black, fill=white!20, inner sep=1pt] (char) {\textbf{S}};
    \end{tikzpicture}  & $-$ & $-$ \\
   NAUTILUS~\cite{xu2025nautilus}  & Image & masks/open-end & Seg./Caption. & 1.5M & $-$ 
    & \crossed & \begin{tikzpicture}[baseline=(char.base)]
        \node[shape=circle, draw=black, fill=orange!50, inner sep=1pt] (char) {\textbf{A}};
    \end{tikzpicture} & \begin{tikzpicture}[baseline=(char.base)]
        \node[shape=circle, draw=black, fill=red!20, inner sep=1pt] (char) {\textbf{S}};
    \end{tikzpicture}  & \begin{tikzpicture}[baseline=(char.base)]
        \node[shape=circle, draw=black, fill=white!20, inner sep=1pt] (char) {\textbf{S}};
    \end{tikzpicture}  & $-$ & $-$  \\
   UTB180 \cite{alawode2022utb180} & Video & masks/bounding-boxes & Tracking & 180 & $-$ 
    & \crossed & $-$ & \begin{tikzpicture}[baseline=(char.base)]
        \node[shape=circle, draw=black, fill=red!20, inner sep=1pt] (char) {\textbf{S}};
    \end{tikzpicture}  & \begin{tikzpicture}[baseline=(char.base)]
        \node[shape=circle, draw=black, fill=gray!30, inner sep=1pt] (char) {\textbf{D}};
    \end{tikzpicture}  & $-$ & $-$ \\
   WebUOT \cite{zhang2024webuot1madvancingdeepunderwater} & Video & masks/bounding-boxes & Tracking & 1M & $-$ 
    & \crossed & $-$ & \begin{tikzpicture}[baseline=(char.base)]
        \node[shape=circle, draw=black, fill=red!20, inner sep=1pt] (char) {\textbf{S}};
    \end{tikzpicture}  & \begin{tikzpicture}[baseline=(char.base)]
        \node[shape=circle, draw=black, fill=gray!30, inner sep=1pt] (char) {\textbf{D}};
    \end{tikzpicture}  & $-$ & $-$ \\
    % Marine Domains \\
    \midrule
    MarineGPT~\cite{zheng2023marinegpt} & Image & open-end & Species Und. & 5M & $-$ & \crossed & \begin{tikzpicture}[baseline=(char.base)]
        \node[shape=circle, draw=black, fill=orange!50, inner sep=1pt] (char) {\textbf{A}};
    \end{tikzpicture} & $-$ & $-$ & $-$ & $-$ \\ MarineEval~\cite{wong2025marineeval}  & Image & multiple-format & General Und. & 2K & 20 & \crossed  & \begin{tikzpicture}[baseline=(char.base)]
        \node[shape=circle, draw=black, fill=orange!50, inner sep=1pt] (char) {\textbf{A}};
    \end{tikzpicture}; \begin{tikzpicture}[baseline=(char.base)]
        \node[shape=circle, draw=black, fill=purple!40, inner sep=1pt] (char) {\textbf{B}};
    \end{tikzpicture} & \begin{tikzpicture}[baseline=(char.base)]
        \node[shape=circle, draw=black, fill=red!20, inner sep=1pt] (char) {\textbf{S}};
    \end{tikzpicture}; \begin{tikzpicture}[baseline=(char.base)]
        \node[shape=circle, draw=black, fill=green!20, inner sep=1pt] (char) {\textbf{E}};
    \end{tikzpicture} & \begin{tikzpicture}[baseline=(char.base)]
        \node[shape=circle, draw=black, fill=white!20, inner sep=1pt] (char) {\textbf{S}};
    \end{tikzpicture}  & $-$ & $-$ \\
    CoralVQA~\cite{han2025coralvqa} & Image & exact-match & Coral Und. & 177K & 16 & \crossed & \begin{tikzpicture}[baseline=(char.base)]
        \node[shape=circle, draw=black, fill=orange!50, inner sep=1pt] (char) {\textbf{A}};
    \end{tikzpicture} & \begin{tikzpicture}[baseline=(char.base)]
        \node[shape=circle, draw=black, fill=red!20, inner sep=1pt] (char) {\textbf{S}};
    \end{tikzpicture} & \begin{tikzpicture}[baseline=(char.base)]
        \node[shape=circle, draw=black, fill=white!10, inner sep=1pt] (char) {\textbf{S}};
    \end{tikzpicture} & $-$ & $-$ \\
    UWBench~\cite{zhang2025uwbenchcomprehensivevisionlanguagebenchmark} & Image & multiple-format & General Und. & 125K & $-$  & \crossed & \begin{tikzpicture}[baseline=(char.base)]
        \node[shape=circle, draw=black, fill=orange!50, inner sep=1pt] (char) {\textbf{A}};
    \end{tikzpicture} & \begin{tikzpicture}[baseline=(char.base)]
        \node[shape=circle, draw=black, fill=red!20, inner sep=1pt] (char) {\textbf{S}};
    \end{tikzpicture}; \begin{tikzpicture}[baseline=(char.base)]
        \node[shape=circle, draw=black, fill=green!20, inner sep=1pt] (char) {\textbf{E}};
    \end{tikzpicture} & \begin{tikzpicture}[baseline=(char.base)]
        \node[shape=circle, draw=black, fill=white!10, inner sep=1pt] (char) {\textbf{S}};
    \end{tikzpicture}  & $-$ & $-$ \\
    UVLM~\cite{xue2025uvlmbenchmarkingvideolanguage} & Video & open-end & General Und. & 2K & 9 &   \crossed & \begin{tikzpicture}[baseline=(char.base)]
        \node[shape=circle, draw=black, fill=orange!50, inner sep=1pt] (char) {\textbf{A}};
    \end{tikzpicture} & \begin{tikzpicture}[baseline=(char.base)]
        \node[shape=circle, draw=black, fill=red!20, inner sep=1pt] (char) {\textbf{S}};
    \end{tikzpicture}; \begin{tikzpicture}[baseline=(char.base)]
        \node[shape=circle, draw=black, fill=green!20, inner sep=1pt] (char) {\textbf{E}};
    \end{tikzpicture} & $-$ & $-$ & $-$ \\ 
    \midrule
    \textbf{Ours} & Video & multiple-format & Event Und.  & 20K & 20 & \checked & \begin{tikzpicture}[baseline=(char.base)]
        \node[shape=circle, draw=black, fill=orange!50, inner sep=1pt] (char) {\textbf{A}};
    \end{tikzpicture}; \begin{tikzpicture}[baseline=(char.base)]
        \node[shape=circle, draw=black, fill=purple!40, inner sep=1pt] (char) {\textbf{B}};
    \end{tikzpicture} & \begin{tikzpicture}[baseline=(char.base)]
        \node[shape=circle, draw=black, fill=red!20, inner sep=1pt] (char) {\textbf{S}};
    \end{tikzpicture}; \begin{tikzpicture}[baseline=(char.base)]
        \node[shape=circle, draw=black, fill=yellow!20, inner sep=1pt] (char) {\textbf{H}};
    \end{tikzpicture}; \begin{tikzpicture}[baseline=(char.base)]
        \node[shape=circle, draw=black, fill=green!20, inner sep=1pt] (char) {\textbf{E}};
    \end{tikzpicture} & \begin{tikzpicture}[baseline=(char.base)]
        \node[shape=circle, draw=black, fill=white!20, inner sep=1pt] (char) {\textbf{S}};
    \end{tikzpicture}; \begin{tikzpicture}[baseline=(char.base)]
        \node[shape=circle, draw=black, fill=gray!30, inner sep=1pt] (char) {\textbf{D}};
    \end{tikzpicture} & \begin{tikzpicture}[baseline=(char.base)]
        \node[shape=circle, draw=black, fill=pink!50, inner sep=1pt] (char) {\textbf{T}};
    \end{tikzpicture}  & \begin{tikzpicture}[baseline=(char.base)]
        \node[shape=circle, draw=black, fill=blue!30, inner sep=1pt] (char) {\textbf{R}};
    \end{tikzpicture}; \begin{tikzpicture}[baseline=(char.base)]
        \node[shape=circle, draw=black, fill=blue!50, inner sep=1pt] (char) {\textbf{O}};
    \end{tikzpicture}   \\ 
    \bottomrule
    \end{tabular}%
}
% \vspace{-5mm}
\end{table}

% \vspace{-0.19in}
\section{Methodology}
% \vspace{-0.12in}
\subsection{MarineEVT Construction}
\label{sec:motivation}
% \vspace{-0.06in}
\subsubsection{Clarification with Existing Datasets and Benchmarks}
Marine videos are visually redundant and unevenly informative, making it hard to localize rare yet significant events and complex behaviors. To bridge this gap, we introduce MarineEVT, an event-centric dataset specifically designed to advance temporal and causal understanding of marine videos. It provides fine-grained, temporally grounded tasks to assess VLM reasoning in dynamic marine environments, going beyond \textit{object‑} or \textit{image‑} level benchmarks to enable deeper temporal and ecological understanding. In Table~\ref{tab:comparisions}, we compare MarineEVT with existing marine datasets and benchmarks. MarineEVT precisely targets event-centric reasoning, encompassing \textit{what}, \textit{which}, \textit{where}, \textit{when}, and \textit{why} of marine events. Meanwhile, it anchors its evaluation in four distinct question types grounded in domain knowledge, thereby assessing models' ability to integrate semantic reasoning, temporal causality, and ecological understanding for autonomous marine observation, environmental monitoring, and scientific discovery. 

% , which struggle to interpret temporally linked, causally structured dynamics, often overlooking subtle cues that distinguish meaningful ecological interactions from routine observations

% \vspace{-5mm}
\subsubsection{MarineEVT Construction Pipeline.}

\begin{figure}[t]
    \centering
   \includegraphics[width=\textwidth]{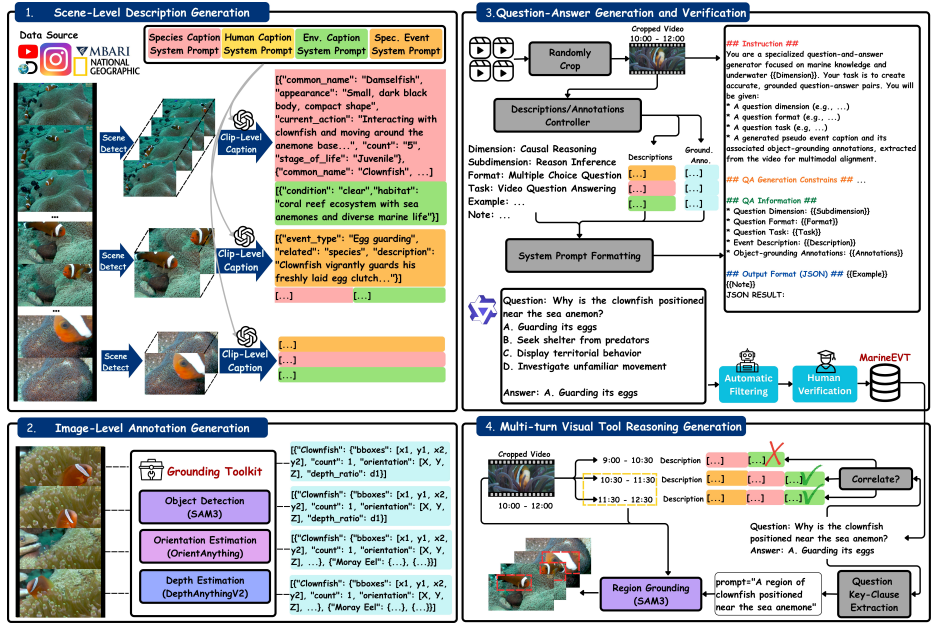}
    % \vspace{-7mm}
   \captionof{figure}{We propose a data construction pipeline that systematically transforms publicly available marine videos into high‑quality, human‑verified annotations, ensuring reliability and versatility for diverse VLM training tasks.}
   \label{fig:construction}
   % \vspace{-7.5mm}
\end{figure}

We propose a scalable pipeline with hierarchical multi-level verification for reliable data construction as shown in Fig.~\ref{fig:construction}. We collect 7,300 marine videos from sources, including MBARI~\cite{mbari}, Discovery Channel~\cite{discovery}, National Geographic~\cite{nationalgeographic} from YouTube~\cite{youtube}, and Instagram~\cite{instagram}. Considering video QA tasks introduce temporal complexities, we adopt a two-stage approach: \textit{scene-level} pseudo-captions for context and \textit{image-level} annotations for fine-grained grounding. During construction, we incorporate marine domain expertise into both \textbf{prompt design} and \textbf{data annotation verification}, ensuring that VLMs are guided by precise ecological terminology and contextually aligned instructions. This expert integration enables the model to better interpret biologically meaningful cues and reduce ambiguity:

\noindent\textbf{Two-stage data generation} from coarse-descriptions to fine-annotation. Directly prompting an LLM to generate question-answer pairs from raw marine video is inevitably unreliable due to the knowledge gap, and the critical information is sparse and unevenly distributed in marine videos. To mitigate these, we adopt a two-stage coarse-to-fine framework. First, to address the intrinsic sparsity of marine events, we employ TransNet~\cite{soucek2019transnet} to segment \textbf{7,300} videos into \textbf{97,284} scenes. For each scene, we use \textbf{domain-specialized prompts}, covering species, humans, environments, and notable events to guide GPT-5~\cite{singh2025openaigpt5card} in generating coarse-grained, domain-specific scene-level descriptions. Second, for each image in the scene, we apply grounding toolkits to extract image-level grounding annotations for each entity driven from the descriptions. Toolkits include SAM3~\cite{carion2025sam3segmentconcepts} for detection, DepthAnythingV2~\cite{depth_anything_v2} for depth estimation, and OrientAnything~\cite{orient_anything} for orientation (totally \textbf{94,028} object-grounding annotations consist of \textbf{240,017} bounding boxes), enriching descriptions with spatial and geometric grounding. These intermediate descriptions and annotations establish a semantic foundation that enables the LLM to subsequently produce accurate, fine-grained question-answer pairs with improved grounding and consistency.

\noindent\textbf{Question-answer generation} with visual tool reasoning steps generation under rigorous verification. Following the previous generation stage, each scene contains up to four domain-specific descriptions and $N$ grounding annotations (where $N$ is the number of frames). To augment data diversity, we randomly crop the original video into sequences containing $k$ sub-scenes, yielding up to $4 \times k$ descriptions and $N \times k$ annotations. Combined with metadata (\emph{e.g.}, question task (yes/no, MCQ, open-end, exact-match), reasoning dimension), these form structured prompts for QA generation via QwenVL-Max~\cite{bai2025qwenvl3}. All generated pairs undergo \textbf{rigorous two-tier verification}: automated filtering by three LLMs, followed by validation from three expert human annotators, before final ingestion into MarineEVT. In addition, we \textbf{synthesize intermediate reasoning steps} to enable VLMs to localize critical information. GPT-5~\cite{singh2025openaigpt5card} validates the correlation between sub-scene descriptions and QA pairs, introducing temporal grounding steps when necessary. Subsequently, SAM3~\cite{carion2025sam3segmentconcepts} generates \textbf{spatial bounding boxes} based on the question's key intent. This process yields data ready for tool-integrated training.

% \begin{itemize}[leftmargin=0pt]
% \item \textbf{Question-Answer generation with rigorous verification}. We sample videos and crop short temporal sequences. For each, we retrieve pseudo scene descriptions, grounding annotations, and metadata (\emph{e.g.}, question-format, dimension) to construct structured prompts for QA generation via QwenVL-Max~\cite{bai2025qwenvl3}. Generated pairs undergo verification by 3 LLMs filtering, then by 3 expert human annotators, before ingestion into MarineEVT.
% \item \textbf{Visual-tool invocation reasoning steps generation}. To train models in tool invocation for pivotal spatiotemporal moments, we construct multi-turn reasoning steps for MarineEVT. For each QA pair, a LLM checks if relevant sub-scene descriptions of crop videos involve the question intention; if yes, we sample around the verified sub-scene. We then extract key semantic clauses to prompt region grounding turns, enabling fine-grained visual alignment during interactive reasoning.
% \end{itemize}
% \vspace{-0.05in}
% This hybrid strategy ensures factual and semantically rich QA pairs.

% \vspace{-0.15in}
\subsection{Multi-Turn Visual Tool-Integrated Reasoning}
% \vspace{-0.06in}
\label{sec:method}
We argue that it is really challenging to localize sparse but critical information within the marine videos, as supported by the above-mentioned challenges. In this work, we propose to integrate the visual tools for assisting event-centric video understanding. Specifically, we decompose a complicated video understanding problem into multiple steps, where in each step we can call corresponding expert models for generating intermediate outputs. Meanwhile, the intermediate outputs can provide extra visual cues and highlight the relevant information regarding user intents. Formally, we formulate the \textbf{multi-turn visual tool-integrated reasoning} as a sequential decision-making process, where a VLM $\pi_{\theta}$, constructs a reasoning trajectory to solve a user-specified task. This trajectory consists of interleaved tool invocations, enabling the model to \textit{dynamically plan}, \textit{gather external evidence}, and \textit{refine its internal representations} for robust and interpretable event-centric video understanding.  Given query $q$, visual input $V_0$, and toolbox $\mathcal{T}$, the VLM interacts over $K$ steps. At step $k$, conditioned on context $(q, V_k, H_{k})$, the model generates reasoning $r_k$, selects tools $\mathcal{T}_k \subseteq \mathcal{T}$, and invoke calls $\{c_{k,j}\}$. 
The calling tools return observations $\mathbf{o}_{k,j}$ that augment previous visuals, and this process continues until termination, yielding the final answer $a$. The trajectory is described as:
\begin{equation}
 \small
    \zeta = \Big( (r_1, \mathcal{T}_1, \{c_{1,j}\}, \{\mathbf{o}_{1,j}\}), \dots, (r_K, \mathcal{T}_K, \{c_{K,j}\}, \{\mathbf{o}_{K,j}\}), a \Big),
\end{equation}
we optimize $\pi_{\theta}$ to maximize answer correctness and cumulative rewards:
\begin{equation}
\label{equa:objective}
\small
\max_{\pi} \mathbb{E}_{\pi} \left[\sum_{n=1}^N \sum_{k=1}^{K} R( \pi_{\theta} (r_k, \mathcal{T}_{k}, \mathbf{o}_{k} \mid q_n,V_{n,k},H_{n,k}), g_{n, k}) \right],
\end{equation}
where $R$ denotes a reward function that quantifies the validity and utility of the $k$-th reasoning step, including logical consistency $r_{k}$, appropriate tool $\mathcal{T}_{k}$, and the groundedness of the resulting observation $o_{k}$ conditioned on the state $k$, which encapsulates the query $q_n$, visual input $V_{n,k}$, and the history $H_{n,k}$.

\begin{figure}[t]
    \centering
   \includegraphics[width=\textwidth]{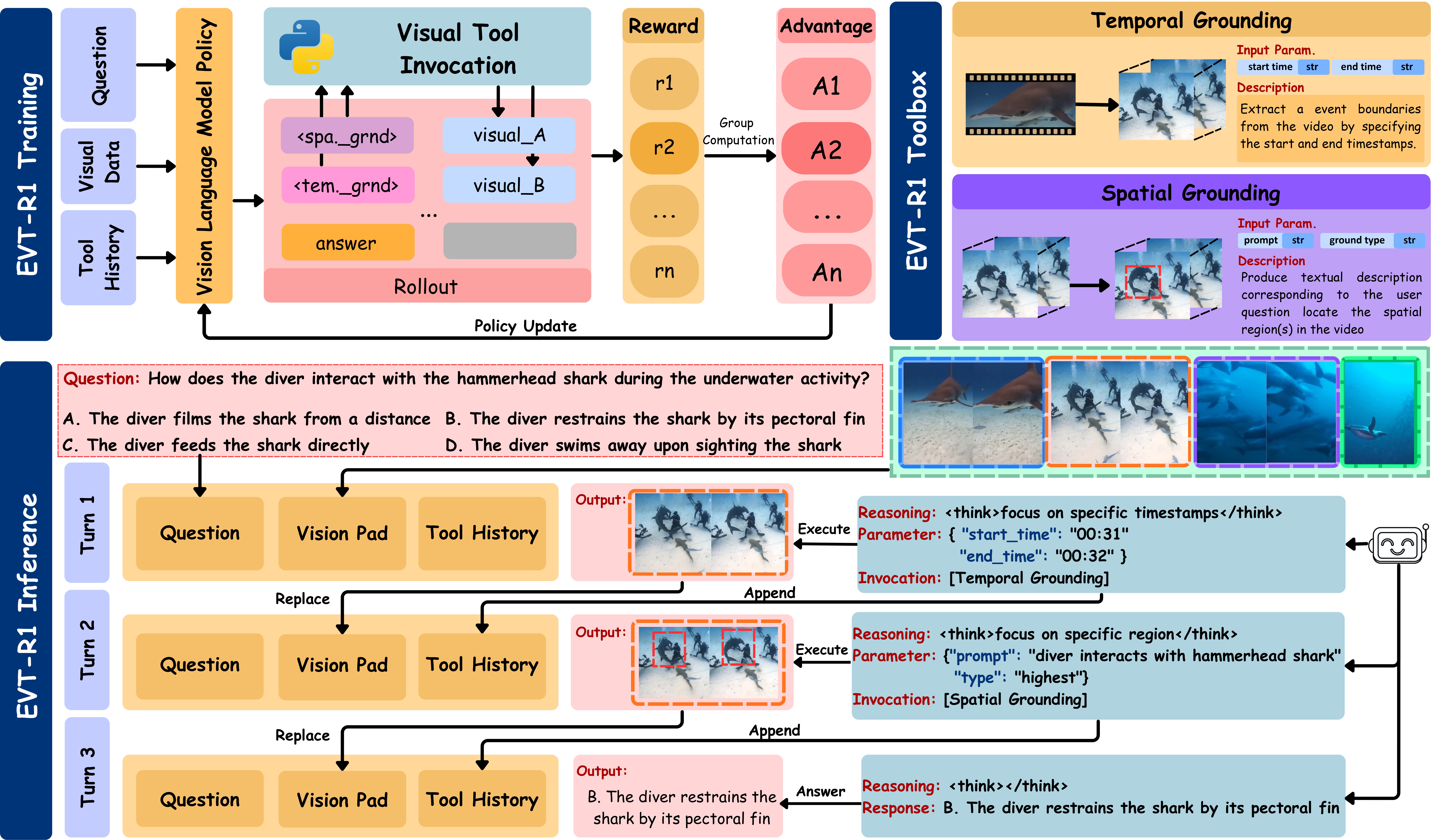}
   % \vspace{-0.2in}
   \captionof{figure}{The training and inference processes of EVT-R1. 
    }
   \label{fig:pipeline}
   % \vspace{-0.25in}
\end{figure}
% \vspace{-0.13in}
\subsection{EVT-R1 Overview}
% \vspace{-0.06in}
The proposed EVT-R1 leverages reinforcement learning (RL) to optimize VLMs for flexible reasoning and strategic visual tool invocation. As illustrated in Fig.~\ref{fig:pipeline}, the policy model accesses temporal and spatial localization tools within toolbox $\mathcal{T}$. During inference, given a user query and video sequence, the policy $\pi_{\theta}$ dynamically decides whether to invoke a tool. Upon invocation, the tool modifies visual inputs (\emph{e.g.}, via temporal or spatial grounding), replacing prior visual inputs in the dialogue history to remove outdated tokens and redundancy, which is essential for VLMs to gather critical and relevant information for generating reliable answers. Then the VLM updates its reasoning over updated visual inputs and tool-call history, stopping once the policy finds the evidence sufficient for a final answer. Meanwhile, during training, the VLM policy produces a group of responses, including tool invocations and new visual outputs, or answers only. These rollouts are evaluated by a \textbf{reward model} to guide VLM backpropagation. The optimization encourages the VLM to either invoke tools or directly output an answer at each turn. Finally, please note that EVT-R1 is different from chain-of-thought~\cite{wei2023chainofthoughtpromptingelicitsreasoning}, which does not involve tool calling and visual updating.

% \vspace{-4mm}
\subsection{Reward Model}
\label{subsec:EVT-R1grpo}
% \vspace{-0.06in}
\begin{figure}[t]
    % \vspace{-2mm}
    \centering
    \scalebox{0.95}{
   \includegraphics[width=\textwidth]{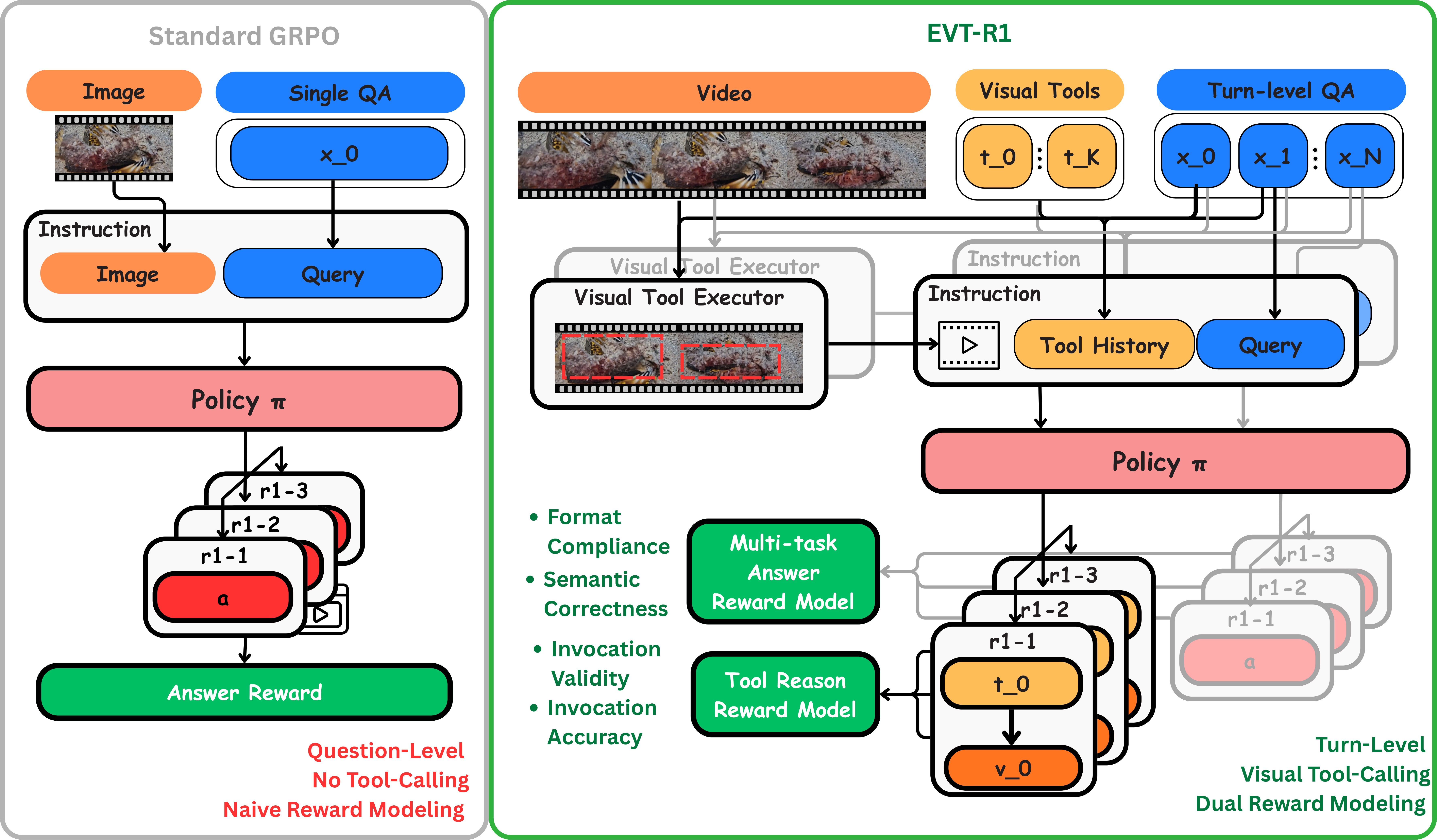}
   }
   % \vspace{-0.12in}
   \captionof{figure}{Compared with GRPO, EVT-R1 devises separate rewards for tool usage and answer accuracy, providing more informative intermediate feedback.} 
   \label{fig:method}
   % \vspace{-0.15in}
\end{figure}

Our reward model is modified from group relative policy optimization (GRPO)~\cite{guo2025grpo}. Differently, unlike existing works~\cite{guo2025grpo,wu2025vtoolr1vlmslearnthink}, which assign rewards only based on the final output, our EVT-R1 devises separated rewards for \textbf{tool usage} and \textbf{answer accuracy} as shown in Fig.~\ref{fig:method}. Our design enables turn-level RL, guiding the model to produce correct answers and learn when and how to use visual tools effectively. Specifically, we propose a dual-component reward model that corrects final answers while explicitly encouraging effective intermediate tool-use decisions. The dual-reward model consists of: \textbf{(i)} a tool-reasoning reward $R_{\text{tool}}$ that assesses whether invoking a tool was valid and accurate at each step, and \textbf{(ii)} a multi-task answer reward $R_{\text{ans}}$ that evaluates the correctness of the final response. Please refer to our Supp. for details:
% \vspace{-0.06in}
\begin{equation}
\label{equa:rewards}
 \small
R(y_i, g_i) = \lambda  \mathbb{I}[\text{tool\_turn}](R_{\text{tool}}(y_i, g_i)) + (1-\lambda)\mathbb{I}\text{[answer\_turn}](R_{\text{ans}}(y_i, g_i)),
\end{equation}
% \vspace{-0.06in}
where we detail $R_{\text{tool}}$ and $R_{\text{ans}}$ as follows:
% \vspace{-0.04in}
\begin{itemize}[leftmargin=0pt]
    \item \textbf{Tool-reasoning reward $R_{\text{tool}}$} evaluates two aspects: \textit{Invocation Validity}, which awards a binary score (1/0) based on whether the tool invocation matches the ground truth at each turn; and \textit{Invocation Accuracy}, an outcome-based metric scoring 1 if the tool's visual output matches the ground truth, and 0 otherwise.
    \item \textbf{Multi-task answer reward $R_{\text{ans}}$} evaluates two criteria: \textit{Format Compliance}, which checks adherence to the expected output format (\emph{e.g.}, JSON structure, long-short answer); and \textit{Semantic Correctness}, computed via exact string matching for closed-form tasks (\emph{e.g.}, yes/no, MCQ, exact-match) or cosine similarity over embeddings for open-ended answers. Scores are normalized per group to down-weight hard-negative, low-similarity responses.
\end{itemize}
% \vspace{-0.04in}
% % More details are included in our Supp regarding the notations and procedures.

\begin{algorithm}[t]
% \small
\caption{\small EVT-R1 Training}
\label{alg:turn_level_grpo}
\begin{algorithmic}[1]
\REQUIRE Initial policy $\pi_{\theta}$, reward $R$, dataset $\mathcal{D}$, group size $G$, clip parameters $\varepsilon$
\ENSURE $\pi_{\theta}$
\FOR{each training iteration}
    \STATE Update old policy: $\pi_{\theta_{\text{old}}} \leftarrow \pi_{\theta}$
    \STATE Sample a batch of input queries $\mathcal{Q}_b \sim \mathcal{D}$
    \FOR{each query $q, V, H  \in \mathcal{Q}_b$}
        \STATE Process V based on the current H step: $V' = process(V, H)$ 
        \STATE Sampling $G$ actions $\{o_t^i\}_{i=1}^G \sim \pi_{\theta_{\text{old}}}(\cdot \mid q, V', H)$
        \STATE Calculate dual rewards $R_{tool}$ or $R_{ans}$ subjected to current turn $g_i$ (Equ~\ref{equa:rewards})
        % \IF {$g_i$ is tool calling response}
        %     \STATE  Compute tool-reason rewards $\{r_t^i\}_{i=1}^G = R_{tool}(\{y_t^i\}, g_i)$
        % \ELSE 
        %     \STATE Compute multi-task answer rewards $\{r_t^i\}_{i=1}^G = R_{ans}(\{y_t^i\}, g_i)$
        % \ENDIF
        \STATE Calculate turn-level advantage $\hat{A}_t^i$ 
    \ENDFOR
    \STATE Collect all turn-level rollouts into one batch
    \STATE Update policy model $\pi_{\theta}$ by maximizing objective $\mathcal{L}_{GRPO}({\theta})$ (Equ~\ref{equa:grpo})
\ENDFOR
\end{algorithmic}
\end{algorithm}

% \vspace{-0.15in}
\subsection{Training Objective and Pseudocode}
\label{subsec:objective}
% \vspace{-0.06in}
Finally, we detail the whole training procedure of EVT-R1 in Algorithm~\ref{alg:turn_level_grpo}. Concretely, given a multimodal input triplet $[q,V,H]$, the algorithm draws several output sequences $\{ o_i \}_{i=1}^G \sim \pi_{\text{old}}(\cdot \mid q, V, H)$ from the current policy, evaluates their relative quality, and updates the policy parameters to favor higher-scoring responses within each group, optimizing the objective $\mathcal{L}_{\text{GRPO}}(\theta)$ as follows:
% \vspace{-0.06in}
\begin{equation}
\label{equa:grpo}
 \small
 \centering
\resizebox{0.9\hsize}{!}{$\displaystyle \mathcal{L}_{\text{GRPO}}(\theta) = \frac{1}{G} \sum_{i=1}^{G} \frac{1}{|o_i|} \sum_{t=1}^{|o_i|} \min\left[ \frac{\pi_{\theta}(o_{i,t}|q,o_{i,<t})}{\pi_{\theta_{\text{old}}}(o_{i,t}|q,o_{i,<t})} \hat{A}_{i,t}, c(\epsilon, \hat{A}_{i,t}) \right] - \beta D_{\text{KL}}[\pi_{\theta}||\pi_{\text{ref}}]$},
% \vspace{-0.06in}
\end{equation}
where $\epsilon$, $\beta$ are hyperparameters, $c$ is the clip advantage, $D_{\text{KL}}$ denotes KL divergence, which is used as a regularization to avoid the unstable update of the new policy $\pi_{\theta}$ compared to the reference policy $\pi_{\text{ref}}$. 
% \vspace{-0.15in}
\section{Experiments}
\label{sec:experiments}
% \vspace{-0.1in}
\subsection{Experimental Setting}
% \vspace{-0.08in}
\noindent\textbf{Baselines}. We evaluate general-purpose and domain-specific video VLMs. Due to limited marine models, we fine-tune open-source VLMs on curated marine data using three strategies: supervised fine-tuning (SFT), GRPO~\cite{guo2025grpo} for post-training only, and our EVT-R1. We leverage Qwen3-VL~\cite{bai2025qwenvl3} as our baseline. We adopt a two-stage fine-tuning strategy: SFT warm-up with one epoch, followed by RL with 4 epochs. For a fair comparison, we also perform SFT and GRPO for 5 epochs. We compare against Video-LLaVA-7B~\cite{lin2024videollavalearningunitedvisual}, LLaVA-NeXT-Video-7B~\cite{zhang2024llavanext-video}, VideoLLaMA3-7B~\cite{damonlpsg2025videollama3}, InternVL3-8B~\cite{zhu2025internvl3}, Qwen3-VL-8B-Instruct~\cite{bai2025qwenvl3}, and closed-source models, Gemini-3.0-Flash\cite{geminiteam2025geminifamilyhighlycapable}, Grok4-1-FR\cite{xai2025grok4}, and GPT-5-Mini\cite{singh2025openaigpt5card}, in both tool-intergated and tool-free settings. \textbf{Training setup}. We use LORA~\cite{hu2021loralowrankadaptationlarge} finetuning with AdamW~\cite{loshchilov2019decoupledweightdecayregularization} (lr=$5\times10^{-6}$, weight decay=$1\times10^{-2}$), micro-batch size 2 for long sequences (up to 50k tokens), optimizing on a single NVIDIA H800 GPU. Decoding uses temperature 0.1 in bf16 precision for consistency. \textbf{Datasets \& metrics}. We construct a testing set consisting of 2,000 QA pairs from our MarineEVT for evaluation only (others are used for training) and report overall accuracy. Please refer to our Supp. for details.
 
\begin{table}[t]
    \centering
    \caption{Experimental comparison between open-source and closed-source VLMs under various settings. ($\dagger$ means post-training only, 
    \protect\colorbox{first}{\rule{0pt}{1.2ex}\rule{1.2ex}{0pt}} 1st, 
    \protect\colorbox{second}{\rule{0pt}{1.2ex}\rule{1.2ex}{0pt}} 2nd, 
    \protect\colorbox{third}{\rule{0pt}{1.2ex}\rule{1.2ex}{0pt}} 3rd).}
    \label{tab:overalltool}
    % \vspace{-3mm}
    \small
    \scalebox{0.85}{
    \begin{tabular}{l|c|c|cccccc}
        \hline
        \multicolumn{9}{c}{Open-source VLMs w/o Tools} \\
        \hline
        \multicolumn{1}{l|}{Model} &
          \multicolumn{1}{c|}{Training} &
          \multicolumn{1}{c|}{Tools} &
          SemR. &
          ConR. &
          SpaR. &
          TemR. &
          CasR. &
          Avg. \\
        \hline
        \multicolumn{1}{l|}{Video-LLaVA-7B} &
          \crossed &
          \crossed &
          24.40 &
          29.00 &
          8.40 &
          5.80 &
          42.00 &
          21.92 \\
        \multicolumn{1}{l|}{LLaVA-NeXT-Video-7B} &
          \crossed &
          \crossed &
          35.40 &
          34.33 &
          9.00 &
          6.75 &
          42.67 &
          25.63 \\
        \multicolumn{1}{l|}{VideoLLaMA3-7B} &
          \crossed &
          \crossed &
          41.00 &
          46.33 &
          5.40 &
          3.50 &
          61.77 &
          31.60 \\
        InternVL3-8B &
          \crossed &
          \crossed &
          53.40 &
          50.33 &
          17.20 &
          10.33 &
          \cellcolor{third}71.77 &
          40.61 \\
        Qwen3-VL-8B-Instruct &
          \crossed &
          \crossed &
          58.20 &
          \cellcolor{third}52.77 &
          22.40 &
          \cellcolor{third}14.00 &
          71.00 &
          \cellcolor{third}43.67 \\
        \hline
        \textcolor{gray}{Avg. across models} &
          $-$ &
          $-$ &
          \textcolor{gray}{42.48} &
          \textcolor{gray}{42.55} &
          \textcolor{gray}{12.48} &
          \textcolor{gray}{8.08} &
          \textcolor{gray}{57.84} &
          \textcolor{gray}{32.69} \\
        \hline \multicolumn{9}{c}{Closed-source VLMs w/o Tools} \\ 
       \hline
            Grok-4-1-FR &
              \crossed &
               \crossed &
              41.20 &
              37.33 &
              17.40 &
              6.25 &
              50.67 &
              30.57 \\
            Gemini-3.0-Flash &
              \crossed &
               \crossed &
              48.20 &
              43.00 &
              \cellcolor{second}27.00 &
              7.75 &
              62.00 &
              37.59 \\
            GPT-5-Mini &
              \crossed &
               \crossed &
              \cellcolor{third}58.40 &
              30.67 &
              \cellcolor{third}22.80 &
              10.00 &
              66.67 &
              37.71 \\
            \hline
            \textcolor{gray}{Avg. across models} &
              $-$ &
              $-$ &
              \textcolor{gray}{49.27} &
              \textcolor{gray}{37.00} &
              \textcolor{gray}{22.40} &
              \textcolor{gray}{8.00} &
              \textcolor{gray}{59.78} &
              \textcolor{gray}{35.29} \\
        \hline \multicolumn{9}{c}{Fine-tuning open-source VLMs w/ Tools} \\ \hline
        Qwen3-VL-8B (GRPO$\dagger$)  & 
        \checked &
          \checked &
          44.60 &
          37.00 &
          20.00 &
          10.75 &
          62.66 &
          35.58 \\
        Qwen3-VL-8B (SFT) &
          \checked &
          \checked &
          \cellcolor{second}61.40 &
          \cellcolor{first}53.33 &
          22.60 &
          \cellcolor{second}15.00 &
          \cellcolor{first}74.00 &
          \cellcolor{second}45.27 \\
        \textbf{Ours (EVT-R1)} &
          \checked &
          \checked &
          \cellcolor{first}65.80 &
          \cellcolor{first}53.33 &
          \cellcolor{first}30.60 &
          \cellcolor{first}20.75 &
          \cellcolor{first}74.00 &
          \cellcolor{first}48.89 \\
        \hline
    \end{tabular}
    }
    \vspace{-0.25in}
\end{table}

% \vspace{-0.20in}
\subsection{Benchmarking SOTAs}
\begin{figure}[t]
    \centering
   \includegraphics[width=\textwidth]{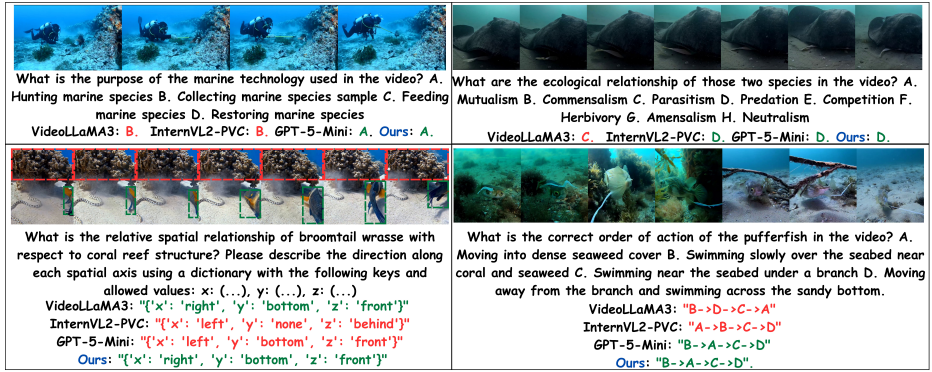}
      % \vspace{-0.25in}
   \captionof{figure}{Experimental result produced by open-source (general-purpose token-compression), closed-source, and our EVT-R1.}
   % \vspace{-0.1in}
   \label{fig:result}
\end{figure}

% \vspace{-0.08in}
\noindent\textbf{Performance analysis}. First, we benchmark \textit{open-source and closed-source VLMs without tool invocation} and \textit{finetuning VLMs with tool invocation} in Table~\ref{tab:overalltool}. Among open-source models, Qwen3-VL-8B-Instruct~\cite{bai2025qwenvl3} outperforms all competitors, achieving the highest average score of 43.67. This superiority is likely attributed to its dynamic-resolution visual encoder, which effectively handles sequences with varying frame sizes. However, a performance gap remains: open-source models underperform in \textit{spatial reasoning} (avg: 12.48) and \textit{temporal reasoning} (8.08). In contrast, while closed-source models also struggle with \textit{temporal reasoning} (8.00), they demonstrate significantly stronger \textit{spatial reasoning} capabilities (avg: 22.40). We provide qualitative comparisons of various algorithms in Fig.~\ref{fig:result}.

Meanwhile, we benchmark \textit{closed-source models with and without tool invocation} as shown in Table~\ref{tab:opensource}. We observe that integrating tools improves the performance of GPT-5-Mini~\cite{singh2025openaigpt5card}, boosting its average score from 37.71 to 40.35. It indicates that an appropriate reasoning process with tool invocation could lead to performance gains without any re-training. In contrast, using the external tools results in performance degradation for the other two closed-source models, likely due to their limited coordination between tool invocation and their internal reasoning. Notably, EVT-R1 surpasses the best closed-source model GPT-5-Mini~\cite{singh2025openaigpt5card} with tool invocation by \textbf{+8.54}, demonstrating the superior performance of our EVT-R1.

\begin{table*}[t]
    \begin{minipage}{.5\linewidth}
        \setlength{\tabcolsep}{4pt}
        \renewcommand{\arraystretch}{1.2}
        \caption{\small Experimental comparisons of closed-source VLMs including GPT-5-Mini~\cite{singh2025openaigpt5card}, Gemini-3.0-Flash~\cite{geminiteam2025geminifamilyhighlycapable}, Grok-4-1-FR\cite{xai2025grok4}: with and without tool invocation.\label{tab:opensource}}
        % \vspace{-0.15in}
        \scalebox{0.55}{
        \begin{tabular}{c|c|ccccccccc}
            \toprule
            Model &
              Tools &
              SemR. &
              ConR. &
              SpaR. &
              TemR. &
              CasR. &
              Avg. \\
            \midrule
             \multirow{2}{*}{Grok-4-1-FR} &
              \crossed &
              41.20 &
              37.33 &
              17.40 &
              6.25 &
              50.67 &
              30.57 \\
             &
              \checked &
              36.00 &
              23.00 &
              13.20 &
              7.75 &
              45.67 &
              25.12$_{\textcolor{red}{-5.45}}$ \\
            \midrule
            \multirow{2}{*}{Gemini-3.0-Flash} &
              \crossed &
              48.20 &
              43.00 &
              \cellcolor{second}{27.00} &
              7.75 &
              62.00 &
              37.59 \\
             &
              \checked &
              47.00 &
              \cellcolor{second}{43.67} &
              25.60 &
              6.25 &
              59.00 &
              36.30$_{\textcolor{red}{-1.29}}$ \\
            \midrule
            \multirow{2}{*}{GPT-5-Mini} &
              \crossed &
              \cellcolor{second}{58.40} &
              30.67 &
              22.80 &
              10.00 &
              \cellcolor{second}{66.67} &
              37.71 \\
             &
              \checked &
              \cellcolor{second}{58.40} &
              \cellcolor{second}{43.67} &
              22.40 &
              \cellcolor{second}{13.00} &
              64.33 &
              \cellcolor{second}{40.35$_{\textcolor{green}{+2.64}}$} \\
              \midrule
                Ours (EVT-R1) &
                \checked &
              \cellcolor{first}{65.80} &
              \cellcolor{first}{53.33} &
              \cellcolor{first}{30.60} &
              \cellcolor{first}{20.75} &
              \cellcolor{first}{74.00} &
              \cellcolor{first}{48.89} \\
              \bottomrule
            \end{tabular}
            }       
    \end{minipage}
    \hfill
    \begin{minipage}{.46\linewidth}
        \centering
        \setlength{\tabcolsep}{4pt}
        \renewcommand{\arraystretch}{1.2}
        % \vspace{-0.1in}
        \caption{\small EVT-R1 vs token compress.
        \label{tab:compress}}
        % \vspace{-0.15in}
        \scalebox{0.47}{
        \begin{tabular}{c|c|cccccc} % Changed to 7 columns total (1 l + 6 c)
        \toprule
        Model & Compress. & SemR. & ConR. & SpaR. & TemR. & CasR. & Avg. \\
        \midrule
        LLaVA-1.5-7B~\cite{liu2023improvedllava} & \multirow{2}{*}{VisionZip~\cite{yang2024visionzip}}  & 37.00 & 18.33 & 10.80 & \cellcolor{second}{6.00} & 46.67 & 23.76 \\ 
        Qwen2.5-VL-7B~\cite{bai2025qwenvl2.5} &  & 26.60 & 13.00 & \cellcolor{second}{12.40} & 2.75 & 47.33 & 20.42 \\ 
        InternVL2-8B~\cite{zhu2025internvl3} & PVC~\cite{yang2025pvc} & \cellcolor{second}{45.60} & \cellcolor{second}{24.67} & 9.40 & 2.50 & \cellcolor{second}{69.00} & \cellcolor{second}{30.23} \\ 
        \midrule
        \textcolor{gray}{Avg. of three} & $-$ & 
              \textcolor{gray}{36.40} &
              \textcolor{gray}{18.67} &
              \textcolor{gray}{10.87} &
              \textcolor{gray}{3.58} &
              \textcolor{gray}{54.33} &
              \textcolor{gray}{24.80} \\
        \midrule
      Ours (EVT-R1) &
        $-$ &
      \cellcolor{first}{65.80} &
      \cellcolor{first}{53.33} &
      \cellcolor{first}{30.60} &
      \cellcolor{first}{20.75} &
      \cellcolor{first}{74.00} &
      \cellcolor{first}{48.89} \\ \bottomrule
    \end{tabular}
    }
    \caption{\small EVT-R1 vs temporal algo.\label{tab:temporal}}
    % \vspace{-4mm}
    \resizebox{\linewidth}{!}{%
        \begin{tabular}{c|cccccc}
        \hline
        % \multicolumn{1}{c|}{\multirow{2}{*}{Model}} &  \multicolumn{6}{c}{Accuracy} \\ \cline{2-7} 
        \multicolumn{1}{c}{Model} & SemR & ConR & SpaR & TemR & CauR & Average \\ \hline
        \multicolumn{7}{l}{\textit{Key-frame selection algorithm}} \\ \hline 
        MaxInfo \cite{li2025maxinfotrainingfreekeyframeselection} & \cellcolor{second}{48.00} & 23.67 & 19.20 & 11.00 & 50.00 & 30.37 \\ 
        AKS ~\cite{tang2025adaptivekeyframesamplinglong} & 45.60 & 20.33 & 17.20 & 12.50 & 48.00 & 28.72 \\ \hline
        \multicolumn{7}{l}{\textit{Frameworks for temporal localization and event-centric video reasoning}} \\ \hline
        VideoITG ~\cite{wang2026videoitgmultimodalvideounderstanding}& 47.40 & 48.67 & \cellcolor{second}{20.20} & \cellcolor{second}{14.25} & 67.33 & \cellcolor{second}{39.57} \\ 
        Chain-of-Frames ~\cite{ghazanfari2026chainofframesadvancingvideounderstanding} & 47.00 & \cellcolor{second}{52.00} & 6.00 & 5.75 & \cellcolor{second}{68.67} & 35.88 \\ \hline
        Ours (EVT-R1) & \cellcolor{first}{65.80} & \cellcolor{first}{53.33} & \cellcolor{first}{30.60} & \cellcolor{first}{20.75} & \cellcolor{first}{74.00} & \cellcolor{first}{48.89} \\ \hline
        \end{tabular}%
        }
    \end{minipage}
% \vspace{-0.2in}
\end{table*}

% \vspace{-0.15in}
\subsection{Ablation Studies}
% \vspace{-0.1in}
\noindent\textbf{Comparison with token compression algorithms} is first included since this line of algorithms is specifically designed to address the redundancy challenge in temporal sequences. We evaluate three open-source VLMs by using two distinct compression methods: VisionZip~\cite{yang2024visionzip} and PVC~\cite{yang2025pvc}. Experimental results are reported in Table~\ref{tab:compress}, which reveals a significant gap between token compression models and our EVT-R1 (\textbf{+18.66} over the best-performing token compression algorithm). We attribute the poor performance of token compression algorithms to their inability to effectively localize critical tokens that convey key information, and in some cases, to the loss of such information during compression.

\noindent\textbf{Comparison with temporal-centric algorithms.} Table~\ref{tab:temporal} compares our method against state-of-the-art key frame selection and temporal localization approaches. EVT-R1 outperforms the best existing baselines by \textbf{+18.52} and \textbf{+9.32}, respectively. These results highlight the inability of current temporal-centric models to adequately capture sparse events across frame and temporal dimensions.

\begin{figure}[t]
    \centering
    \includegraphics[width=\textwidth]{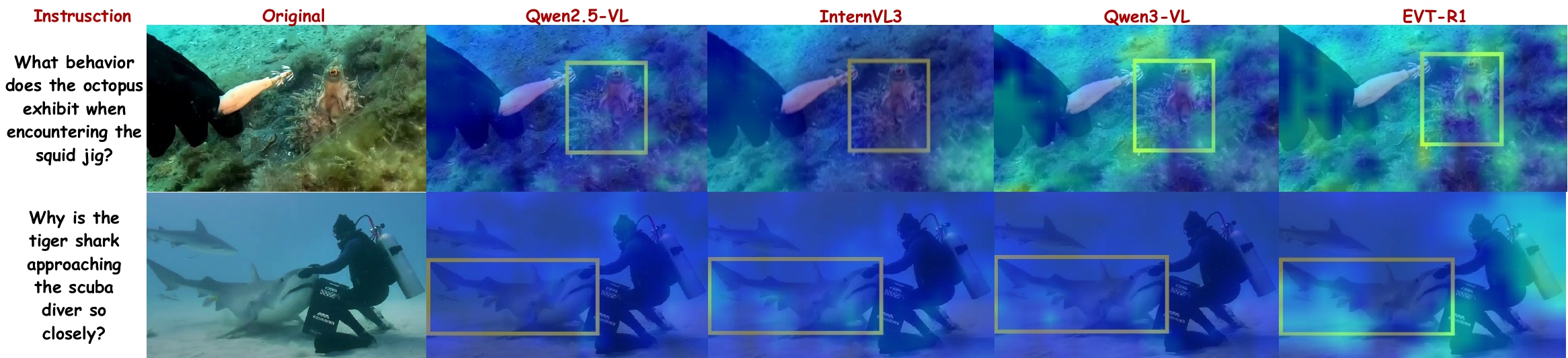}
    % \vspace{-7mm}
    \caption{Attention activation visualization of various models produced by TAM~\cite{tam2025iccv}.}\label{fig:activation}
    % \vspace{-0.18in}
\end{figure}

\begin{table}[t]
    \begin{minipage}{.64\linewidth}
        \centering
       \captionof{table}{\small Ablation studies on tool-calling behavior and spatio-temporal accuracy on testing set.\label{tab:toolbehave}}
        % \vspace{-2mm}
        \resizebox{\linewidth}{!}{%
        \begin{tabular}{l|cc|cc}
        \hline
        \multicolumn{1}{c|}{\multirow{2}{*}{\textbf{Method}}}   & \multicolumn{2}{c}{\textbf{Invocation Validity}} & \multicolumn{2}{|c}{\textbf{Invocation Accuracy}} 
        \\ \cline{2-5}
        & \textbf{Correct Tool} & \textbf{Correct Step} & \multicolumn{1}{|c}{\textbf{Spatial (IoU$\ge$0.5)}} & \textbf{Temporal (IoU$\ge$0.5)} \\ \hline
        Qwen3-VL (SFT-only)  & 65.55 & 84.40 & 80.09 & 14.88 \\
        EVT-R1 (SFT+RL) & \cellcolor{first}{70.00\textcolor{green}{(+4.45)}} & \cellcolor{first}{87.35\textcolor{green}{(+2.95)}} & \cellcolor{first}{83.84\textcolor{green}{(+3.75)}} & \cellcolor{first}{15.00\textcolor{green}{(+0.12)}}\\
        \hline
        \end{tabular}%
    }
        % \includegraphics[width=\textwidth]{images/overlap.pdf}
        % \vspace{-0.17in}
        % \captionof{figure}{\small Average accuracy of various models on spatial and temporal localization over 100 questions.\label{fig:activationpercent}}
    \end{minipage}
    \hfill
    \begin{minipage}{.35\linewidth}
        \centering
        \captionof{table}{\small Average accuracy ($\text{Mean}_{\textcolor{blue}{\text{std}}}$) of 3 trials.\label{tab:trial}}
        \vspace{-0.15in}
        \scalebox{0.66}{
        \begin{tabular}{c|cc}
            \hline
            Question Task  & Blank  & Adversarial \\
            \hline
            Open-end & $0.00_{\textcolor{blue}{0.00}}$ & $0.00_{\textcolor{blue}{0.00}}$ \\
            MCQ & $43.62_{\textcolor{blue}{0.67}}$ & $68.86_{\textcolor{blue}{0.44}}$ \\
            Yes-No & $40.33_{\textcolor{blue}{1.55}}$ & $65.67_{\textcolor{blue}{1.68}}$ \\
            Exact-Match & $1.48_{\textcolor{blue}{0.03}}$ & $10.04_{\textcolor{blue}{0.85}}$ \\
            \hline
            Avg. of tasks & $21.36_{\textcolor{blue}{0.75}}$ & $36.14_{\textcolor{blue}{0.99}}$ \\
            \hline
        \end{tabular}
        }        
    \end{minipage}
    % \vspace{-0.32in}
\end{table}

% \vspace{-0.05in}
\noindent\textbf{Does RL training improve tool-use behavior?} Table~\ref{tab:toolbehave} evaluates whether EVT-R1 learns better tool-calling behavior. RL training significantly improves the model's accuracy in selecting the correct tools at the right steps, yielding gains of \textbf{+3.75} in spatial IoU and \textbf{+0.12} in temporal IoU. Additionally, TAM-generated~\cite{tam2025iccv} attention maps, shown in Fig.~\ref{fig:activation}, reveal that this improved spatial localization enables the model to attend more precisely to critical frame tokens.

% We then conduct experiments to test whether EVT-R1 could achieve accurate spatial localization and temporal event localization. We evaluate spatial accuracy by measuring the overlap between TAM-generated~\cite{tam2025iccv} attention maps and ground-truth interest of regions in Fig.~\ref{fig:activation}, and temporal accuracy by aligning predicted time duration with ground-truth timestamps. EVT-R1 achieves accurate temporal localization (\textbf{+15.0}) and precise spatial localization \textbf{+2.34} as shown in Fig.~\ref{fig:activationpercent}. We noticed that the other three VLMs yield a zero accuracy for temporal localization. This could stem from the end-to-end optimization of these VLMs, which does not explicitly encourage the localization of critical events throughout the whole sequence.

\noindent\textbf{Does visual input matter?} Existing analysis~\cite{chen2024rightwayevaluatinglarge} pointed out that VLMs may directly discard visual inputs and yield responses based on language priors. We conduct similar analysis on MarineEVT under two settings: blank or semantically meaningless inputs and adversarial inputs with temporal/semantic inversions, evaluating whether models rely on visual evidence or default to language priors when cues are absent or misleading. To avoid contamination, we directly use the best-performing commercial VLM GPT-5-Mini~\cite{singh2025openaigpt5card} to do the evaluation since EVT-R1 was optimized on MarineEVT. We observe consistently poor performance in Table~\ref{tab:trial}. Crucially, performance on open-ended/exact-match tasks remains low under adversarial inputs. These results reveal that GPT-5-Mini overly relies on language priors and thus struggles on MarineEVT, highlighting the necessity of VLMs to extract visual cues for answer generation.

\noindent\textbf{Why post-training only does not work}. In Table~\ref{tab:overalltool}, utilizing GRPO for post-training only leads to degraded performance. We manually verified the model outputs and found instability and reward overfitting on long, multi-turn tasks, frequently triggering \textbf{infinite reasoning loops}. It may be caused by the lack of domain knowledge, leading to a weak ability to discriminate when to perform tool invocation or yield final answer. Thus, we first fine-tune the VLM on our MarineEVT for one epoch to alleviate the knowledge gap, followed by GRPO for RL training with a cold start. Such a training strategy leads to an observable performance gain as shown in Table~\ref{tab:grpovsEVT-R1}, revealing that the SFT is an essential step for adapting general-purpose VLMs to specific domains, analogous to that students should have basic knowledge to determine their learning actions.

\begin{table}[t]
    \begin{minipage}{0.49\textwidth}
        \centering
        \includegraphics[width=0.95\textwidth]{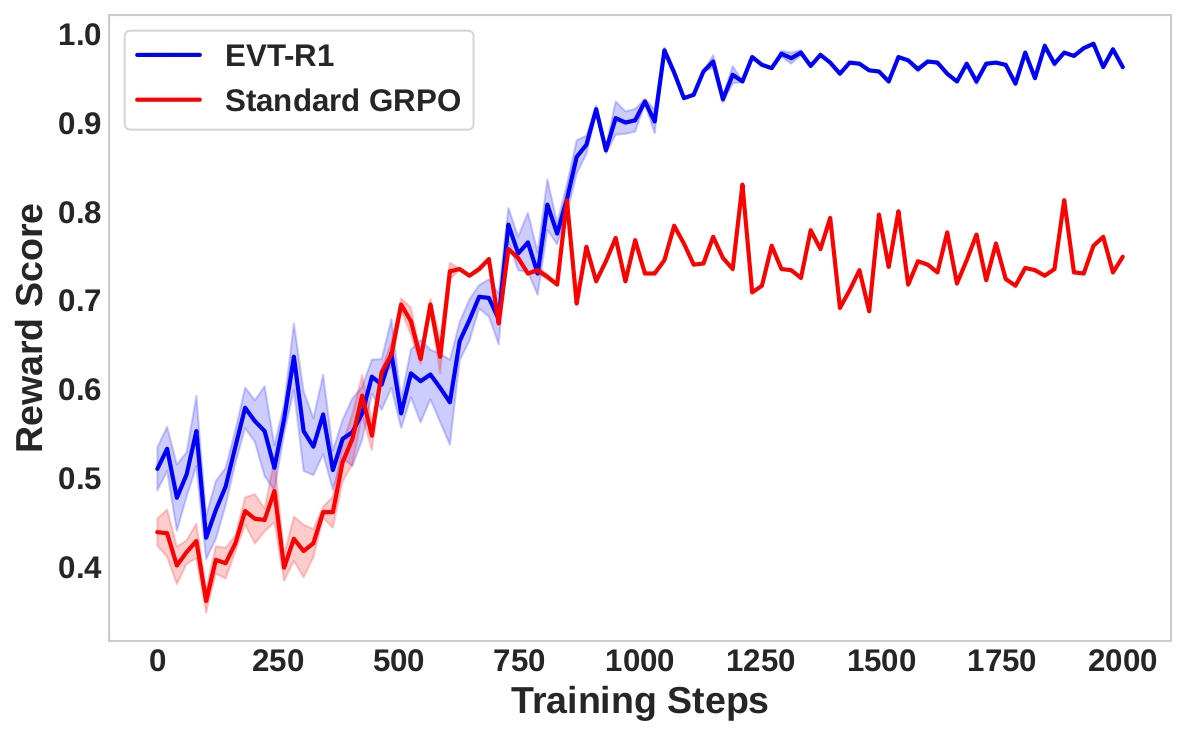}
        % \vspace{-0.0in}
        \captionsetup{type=figure}
        \captionof{figure}{\small We compare the reward score curves of GRPO and EVT-R1.}
        \label{fig:rewardchart}
    \end{minipage}
    \hfill
    \begin{minipage}{0.48\textwidth}
     \centering
        % \vspace{-0.1in}
        \begin{minipage}{\textwidth}
        \caption{\small GRPO vs. EVT-R1.\label{tab:grpovsEVT-R1}}
        % \vspace{-0.15in}
        \resizebox{0.98\textwidth}{!}{%
        \begin{tabular}{l|c|cccccc}
            \hline
            Method & Setting  & SemR. & ConR. & SpaR. & TemR. & CasR. & Avg. \\
            \hline
            GRPO & \multirow{2}{*}{RL only} &
              44.60 &
              37.00 &
              20.00 &
              10.75 &
              62.66 &
              35.58 \\
            EVT-R1 &  &   
                44.80 &
              39.00 &
              21.40 &
              13.75 &
              63.33 &
              36.52 \\
            \hline
             GRPO & \multirow{2}{*}{SFT+RL}  &
              \cellcolor{second}{63.40} &
               \cellcolor{second}{53.33} &
               \cellcolor{second}{29.60} &
               \cellcolor{second}{18.00} &
               \cellcolor{first}{74.67} &
               \cellcolor{second}{48.20} \\ 
            EVT-R1 &  &  
               \cellcolor{first}{65.80} &
              \cellcolor{first}{53.33} &
              \cellcolor{first}{30.60} &
              \cellcolor{first}{20.75} &
              \cellcolor{second}{74.00} &
              \cellcolor{first}{48.89} \\ 
            \hline
        \end{tabular}
        }
        \end{minipage}
        \hfill
        \begin{minipage}{\textwidth}
        \caption{\small Different $\lambda$ coefficient values.}
        \label{tab:coeff}
        % \vspace{-0.15in}
        \resizebox{0.98\textwidth}{!}{%
        \begin{tabular}{l|cccccc}
            \hline
            Coefficient & SemR. & ConR. & SpaR. & TemR. & CasR. & Avg. \\
            \hline
            $\lambda = 0.00$ & 44.80 & \cellcolor{second}{38.67} & 18.80 & \cellcolor{first}{14.50} & 65.00 & 36.35 \\
            $\lambda = 0.25$ & 54.40 & 37.33 & \cellcolor{second}{19.40} & 12.50 & 66.33 & \cellcolor{second}{37.99} \\
            $\lambda = 0.50$ & 54.00 & 38.33 & 18.20 & 12.25 & 66.67 & 37.89 \\
            $\lambda = 0.75$ & \cellcolor{first}{55.40} & \cellcolor{first}{41.33} & \cellcolor{first}{20.80} & 13.50 & \cellcolor{first}{68.33} & \cellcolor{first}{39.87} \\
            $\lambda = 1.00$ & \cellcolor{second}{55.00} & 37.00 & 13.80 & \cellcolor{second}{14.00} & \cellcolor{second}{67.00} & 37.36 \\
            \hline
        \end{tabular}
        }
        \end{minipage}
    \end{minipage}
    \vspace{-0.28in}
\end{table}

\noindent\textbf{Comparison with GRPO}. Under the same experimental settings: RL only and SFT+RL, our EVT-R1 outperforms GRPO by \textbf{+0.94} and \textbf{+0.69}, respectively, demonstrating the effectiveness of the proposed dual-reward design. Meanwhile, we provide the reward curve of GRPO and EVT-R1 in Fig.~\ref{fig:rewardchart}. Our reward function produces higher rewards and more stable convergence compared with GRPO, \emph{e.g.}, a lower variance at later stages (steps 1000–2000), reflecting stable policy updates and effective reward assignment. Finally, we ablate the reward coefficient $\lambda$ in Table~\ref{tab:coeff}. $\lambda = 0.75$ achieves the best performance (avg.~39.87). % Both $\lambda=0.00$ and $\lambda=1.00$ degrade learning by neglecting or overemphasizing reasoning steps, confirming that $\lambda = 0.75$ achieves the right balance between reasoning quality and answer accuracy. 

% In summary, EVT-R1 demonstrates a stronger ability to localize informative spatio-temporal representations for aiding internal reasoning and yielding reliable predictions. 
% \vspace{-6mm}
\section{Conclusion and Acknowledgment}
% \vspace{-3mm}
% \noindent\textbf{Potential applications}. Event-centric marine video understanding addresses a critical bottleneck in marine video analysis: the overwhelming volume of unstructured, repetitive footage where meaningful events, \emph{e.g.}, species traits, or predator–prey interactions typically occupy less than \textbf{1\%} of the whole video. The introduction of MarineEVT promotes developing models to (i) detect semantic or unusual events. (\emph{e.g.}, anomalous motion or behavioral shifts); (ii) characterize the nature of the event (\emph{e.g.}, identifying it as spawning, predation, or pollution discharge); and (iii) summarize and visualize these events to enable efficient human review and downstream scientific analysis, together achieving automatically or semi-automatically educational and ecological analysis.

 % pipelines to detect, characterize, and summarize ecologically significant marine events for

\noindent\textbf{Conclusion}. In this work, we have introduced the first event-centric marine video understanding dataset, MarineEVT, highlighting the specific and intrinsic challenges of marine video: the deep domain expertise requirement and the difficulty to localize and understand the sparse, unpredictable, and unevenly distributed marine events. Besides MarineEVT, we also introduce EVT-R1, the first event-centric visual tool-integrated reasoning framework, where we decompose the video understanding task into the multi-turn tool-integrated reasoning. EVT-R1 demonstrates a stronger ability to localize critical information spatially and temporally than existing algorithms. It also introduces a new direction of using visual tools for the complicated video understanding tasks.

\noindent\textbf{Acknowledgement}. This project was partially supported by Bridging Horizons: An AI-Powered STEM Learning Initiative in Space and Marine Education under the EdUHK–HKUST Joint Centre for Artificial Intelligence and the Marine Conservation Enhancement Fund MCEF22112, and an internal grant frome HKUST (R9429).
We would also like to express our sincere gratitude to the “Sustainable Smart Campus as a Living Lab” (SSC) program at HKUST for its vital support. The program and its dedicated staff not only contributed essential funding and coordination but also fostered the integration of sustainability into campus operations, providing a real-world demonstration of the principles that underpin this research.

\bibliographystyle{splncs04}
\bibliography{main}

% \newpage
% \input{supplementary/statistics}
% \input{supplementary/dimension_detail}
% \input{supplementary/construction}
% \input{supplementary/agentic}

\end{document}

\title{Supplementary Material for \\ MarineEVT: Advancing Event-Centric Marine Video Understanding via Visual Tool Reasoning }

\author{Tuan-An To\inst{1}\orcidlink{0000-0002-5408-8138} \and
Yuk-Kwan Wong\inst{1}\orcidlink{0009-0007-4444-5502} \and
Tuan-Anh Vu\inst{2}\orcidlink{0000-0002-8872-0875}
\and \\
Ziqiang Zheng$^{\dagger}$\inst{1,3}\orcidlink{0000-0002-1477-6040}
\and 
Sai-Kit Yeung\inst{1}\orcidlink{0000-0001-7974-0607}
% \\
% \small
% Project website: \url{https://marineevt.hkustvgd.com}
}

% TODO FINAL: Replace with an abbreviated list of authors.
\authorrunning{A. To et al.}
% First names are abbreviated in the running head.
% If there are more than two authors, 'et al.' is used.

% TODO FINAL: Replace with your institution list.
\institute{The Hong Kong University of Science and Technology, Hong Kong, China \and
University of California, Los Angeles
CA, USA \and
University of Electronic Science and Technology of China, China} 

\titlerunning{MarineEVT}

\maketitle

% \clearpage  % TODO REVIEW/FINAL: This \clearpage needs to be removed from both review and camera-ready versions.
\input{supplementary/event}
\input{supplementary/statistics}
\input{supplementary/construction}
\input{supplementary/dimension}
\input{supplementary/agentic}
\input{supplementary/additional_experiment}
\input{supplementary/discussion}
% ---- Bibliography ----
%
% BibTeX users should specify bibliography style 'splncs04'.
% References will then be sorted and formatted in the correct style.
%
% \newpage
\bibliographystyle{splncs04}
\bibliography{main}